\documentclass[conference]{IEEEtran}
\IEEEoverridecommandlockouts
\usepackage{amsmath,amssymb,amsfonts}
\usepackage{algorithmic}
\usepackage{graphicx}
\usepackage{textcomp}
\usepackage{xcolor}

\usepackage{booktabs} 

\usepackage{kotex}
\graphicspath{{./figures/}}
\usepackage{subcaption}
\usepackage[ruled]{algorithm2e}
\usepackage{caption}
\usepackage{color}
\usepackage{ctable}
\usepackage{siunitx}
\usepackage{enumitem}
\usepackage{multirow}
\usepackage{balance}

\begin{document}

\title{T2FSNN: Deep Spiking Neural Networks with Time-to-first-spike Coding\\
\thanks{\textsuperscript{*}corresponding author: Sungroh Yoon (sryoon@snu.ac.kr)}
}

\author{\IEEEauthorblockN{Seongsik Park, Seijoon Kim, Byunggook Na, Sungroh Yoon\textsuperscript{*}}
\IEEEauthorblockA{
Department of Electrical and Computer Engineering, ASRI, INMC, and Institute of Engineering Research \\
Seoul National University, Seoul 08826, South Korea
}
}

\maketitle

\begin{abstract}
Spiking neural networks (SNNs) have gained considerable interest due to their energy-efficient characteristics, yet lack of a scalable training algorithm has restricted their applicability in practical machine learning problems.
The deep neural network-to-SNN conversion approach has been widely studied to broaden the applicability of SNNs. 
Most previous studies, however, have not fully utilized spatio-temporal aspects of SNNs, which has led to inefficiency in terms of number of spikes and inference latency. 
In this paper, we present T2FSNN, which introduces the concept of time-to-first-spike coding into deep SNNs using the kernel-based dynamic threshold and dendrite to overcome the aforementioned drawback. 
In addition, we propose gradient-based optimization and early firing methods to further increase the efficiency of the T2FSNN.
According to our results, the proposed methods can reduce inference latency and number of spikes to 22\% and less than 1\%, compared to those of burst coding, which is the state-of-the-art result on the CIFAR-100.

\end{abstract}


\section{Introduction}
Recent advances in deep neural networks (DNNs) have led to state-of-the-art results in a variety of applications. 
These DNNs, however, demand a substantial amount of computation and power consumption, which restricts the employment of deep learning to edge devices.
To overcome these challenges, many studies have focused on reducing the model size and the amount of computation through quantization~\cite{park2018quantized} and pruning~\cite{han2015learning}.
Despite these attempts, the complexity of the DNN models has increased rapidly, making it difficult to deploy the DNNs efficiently in resource-constrained environments, such as mobile devices~\cite{tan2019efficientnet}.

Spiking neural networks (SNNs) have emerged as the next generation of neural networks for their superior energy efficiency caused by the features of integrate-and-fire and event-based operations~\cite{maass1997networks}.
Although SNNs have the potential for improving energy efficiency of artificial neural networks, deep SNNs have not been widely used in many applications due to lack of scalable training algorithms.
Deep SNNs have been struggling to achieve competitive results compared to DNNs.
Although many studies about direct training with approximate stochastic gradient descent (SGD) have been published recently~\cite{jin2018hybrid,wu2019direct}, they have shown unsatisfactory results.

DNN-to-SNN conversion methods have been proposed in recent years~\cite{diehl2015fast,rueckauer2017conversion,kim2019spiking,park2019fast,kim2018deep,zhang2019tdsnn,rueckauer2018conversion} to address the issue of training deep SNNs.
Using the conversion methods, deep SNNs can exploit the DNNs' training performance with the pretrained synaptic weights of DNNs, which results in the competitive performance of SNNs.
There are several factors affecting the efficiency of deep SNNs, including neuron types and neural coding schemes, during the conversion.
Among these factors, many studies have focused on neural coding scheme, which is critical in transmitting accurate information with low energy consumption and latency~\cite{park2019fast,kim2018deep,zhang2019tdsnn,rueckauer2018conversion}.

To date, various neural coding schemes have been applied to SNNs, such as rate~\cite{adrian1926impulses} and temporal coding~\cite{park2019fast,butts2007temporal,montemurro2008phase,thorpe2001spike}.
Rate coding, which is a well-known and commonly used neural coding, utilizes firing rate to represent information~\cite{adrian1926impulses}. 
Many DNN-to-SNN conversion methods have adopted rate coding for its simple implementation and robustness~\cite{diehl2015fast,rueckauer2017conversion,kim2019spiking}. 
However rate coding generates a large number of spikes and has a slow information transmission.
In the wake of rate coding, many temporal coding schemes, including phase~\cite{montemurro2008phase} and burst~\cite{park2019fast}, were introduced to deep SNNs to utilize the temporal information in the spike trains as in biological neural systems.
Deep SNNs with phase coding~\cite{kim2018deep} and burst coding~\cite{park2019fast} have improved the efficiency of inference with comparable results to those of DNNs.
However, they still fall short of the desired performance in terms of latency and number of spikes.

To fully utilize temporal coding, time-to-first-spike (TTFS) coding has been applied to deep SNNs recently~\cite{zhang2019tdsnn,rueckauer2018conversion}.
The study~\cite{rueckauer2018conversion}, where the important information was delivered first, successfully introduced TTFS coding into deep SNNs, leading to a significant reduction in number of spikes.
However, this research did not show satisfactory results.
The TDSNN~\cite{zhang2019tdsnn} with reverse coding, which is a kind of TTFS coding, was proposed to improve the accuracy of deep SNNs.
The deep SNNs with reverse coding achieved competitive results with DNNs.
However, the TDSNN does not report the number of spikes and latency, which are critical measures to evaluate the efficiency of deep SNNs.
Furthermore, the additional spikes from auxiliary neurons and the reverse coding prohibited the deep SNNs from improving inference efficiency.

In this paper, we propose a novel deep SNN model, called T2FSNN, for efficient implementation of TTFS coding in deep SNNs.
The T2FSNN exploits a kernel-based dynamic threshold and dendrite with TTFS coding, where the earlier spikes represent more critical information.
In addition, we propose a gradient descent algorithm to optimize the kernels in the dynamic threshold and dendrite to further improve information transmission efficiency.
Besides, we introduce an early firing method to further reduce inference latency.
To validate the proposed approach, we conducted extensive experiments on the MNIST, CIFAR-10, and CIFAR-100, measuring the accuracy, number of spikes, and latency.
According to the results, we were able to increase the accuracy as well as reduce latency and number of spikes, by applying the proposed methods to the T2FSNN.
Our contributions can be summarized as follows:
\begin{itemize}[topsep=0pt,itemsep=0ex,partopsep=1ex,parsep=1ex,leftmargin=*]
\item \textbf{T2FSNN:} We propose T2FSNN, which is an SNN model with a dynamic threshold and dendrite for TTFS coding to reduce the number of spikes and latency of inference.
\item \textbf{Gradient-based optimization:} We propose a gradient-based optimization approach for improving the efficiency of the T2FSNN.
\item \textbf{Early firing:} We propose an early firing method to further reduce the inference latency of T2FSNN.
\end{itemize}

\section{Background and Related Work}
\subsection{Spiking Neural Networks}

The main difference between SNNs and DNNs lies in how information is transmitted between neurons.
SNNs transmit information through spike trains which contain binary spikes (discrete) rather real values (continuous).
The spike train $S_{i}^{l}(t)$ of $i$th neuron in $l$th layer can be represented as
\vspace{-0.3em}
\begin{equation}
\label{eq:spike_train}
S_{i}^{l}(t) = \sum\nolimits_{t_{i}^{l,(f)} \in F_{i}^{l}}{\delta(t-t_{i}^{l,(f)})} \textrm{,}
\vspace{-0.4em}
\end{equation}
where $\delta(t)$ is the Dirac delta function, $f$ is an index of spike, and $F_{i}^{l}$ is a set of spike times satisfying firing condition which is stated as:
\vspace{-0.4em}
\begin{equation}
\label{eq:firing_time}
t_{i}^{l,(f)}: u_{i}^{l}(t_{i}^{l,(f)}) \geq \theta_{i}^{l}(t_{i}^{l,(f)}) \textrm{,}
\vspace{-0.4em}
\end{equation}
where $u_{i}^{l}(t)$ is a membrane potential and $\theta_{i}^{l}(t)$ is a threshold at time $t$.
Due to this information transmission based on the discrete spikes, SNNs have a feature of event-driven operation, which leads to improving computational energy efficiency.
Please note that we will use $t_{i}^{l}$ instead of $t_{i}^{l,(f)}$ in the rest of this paper for simplicity because each neuron generates only one spike in TTFS coding.

In an integrate-and-fire (IF) neuron, which is one of the widely used neuron types in SNNs, a synaptic input is integrated into a membrane potential as follows:
\vspace{-0.3em}
\begin{equation}
\label{eq:vmem}
    u_{j}^{l}(t) = u_{j}^{l}(t-1) + z_{j}^{l}(t) \textrm{,}
\vspace{-0.4em}
\end{equation}
where $z_{j}^{l}$ is a sum of postsynaptic potential (PSP), which can be described as:
\vspace{-0.4em}
\begin{equation}
\label{eq:psp}
z_{j}^{l}(t) = \sum\nolimits_{i}{w_{ij}^{l} d_{j}^{l}(t)  S_{i}^{l\textrm{-}1}(t)+b_{j}^{l}} \textrm{,}
\vspace{-0.4em}
\end{equation}
where $w_{ij}^{l}$ is a synaptic weight, $d_{j}^{l}$ is a dendrite, and $b_{j}^{l}$ is a bias.
Only the neurons satisfying the firing condition (Eq.~\ref{eq:firing_time}) generate spikes, which increases the sparsity of the spikes.

\subsection{Neural Coding}
\begin{figure}[tbp]
    \centering
    \includegraphics[width=0.98\linewidth]{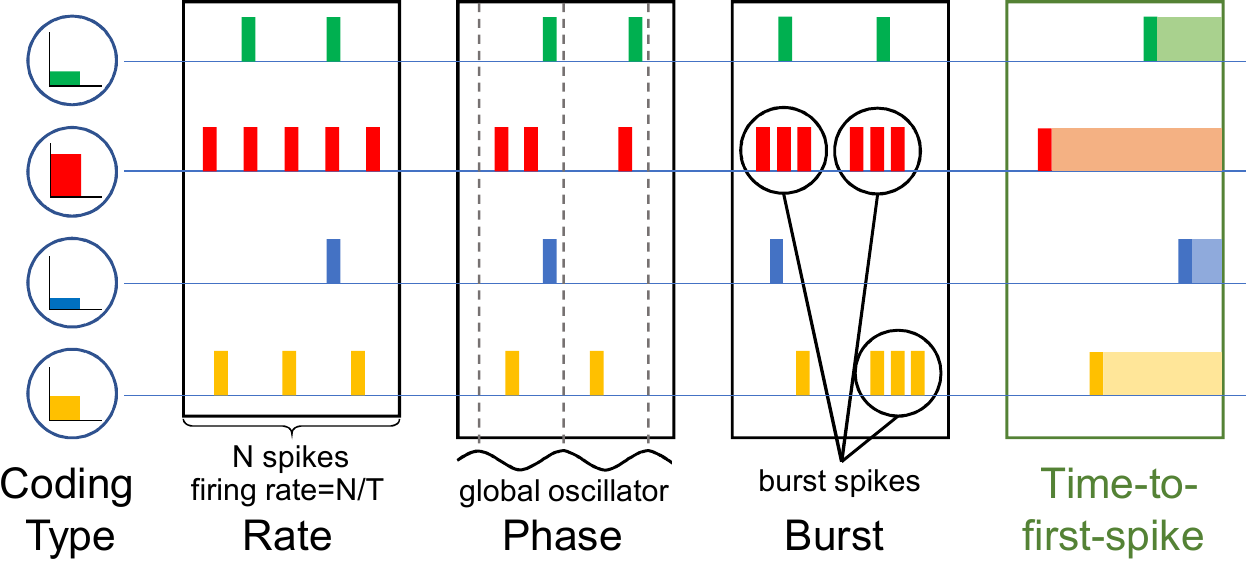}
	\vspace{-0.5em}
	\caption{Various neural coding methods}
	\label{fig:neural_coding}
	\vspace{-2.0em}
\end{figure}

Neural coding is a method of representing information with spike trains, including encoding and decoding procedures.
There have been four neural coding schemes in deep SNNs: rate~\cite{diehl2015fast,rueckauer2017conversion,kim2019spiking}, phase~\cite{kim2018deep}, burst~\cite{park2019fast}, and TTFS~\cite{zhang2019tdsnn,rueckauer2018conversion}, as depicted in Fig.~\ref{fig:neural_coding}.
Rate coding has the advantages of simple implementation and robustness to errors by using firing rate, $\frac{N}{T}$, where $N$ is the total number of spikes in a given time window $T$~\cite{diehl2015fast,rueckauer2017conversion,kim2019spiking,adrian1926impulses}.
However, rate coding cannot utilize temporal information in spike trains and generates a large number of spikes, leading to high energy consumption and long inference latency.

Phase coding encodes temporal information into spike patterns based on a global oscillator~\cite{montemurro2008phase}, and it can significantly reduce the number of spikes in deep SNNs~\cite{kim2018deep}.
However, the efficiency cannot be guaranteed if the input changes dynamically and is unpredictable as hidden layers in deep SNNs~\cite{park2019fast}.
Burst coding attempts to overcome this challenge by introducing burst spikes utilizing inter-spike interval~\cite{park2019fast}.
Burst spikes can carry more information quickly and accurately by inducing PSP dramatically.
Burst coding was able to significantly reduce the number of spikes and improve overall performance; however, it is still short of desired performance in terms of latency and efficiency.

\begin{figure}[tbp]
    \centering
    \includegraphics[width=0.95\linewidth]{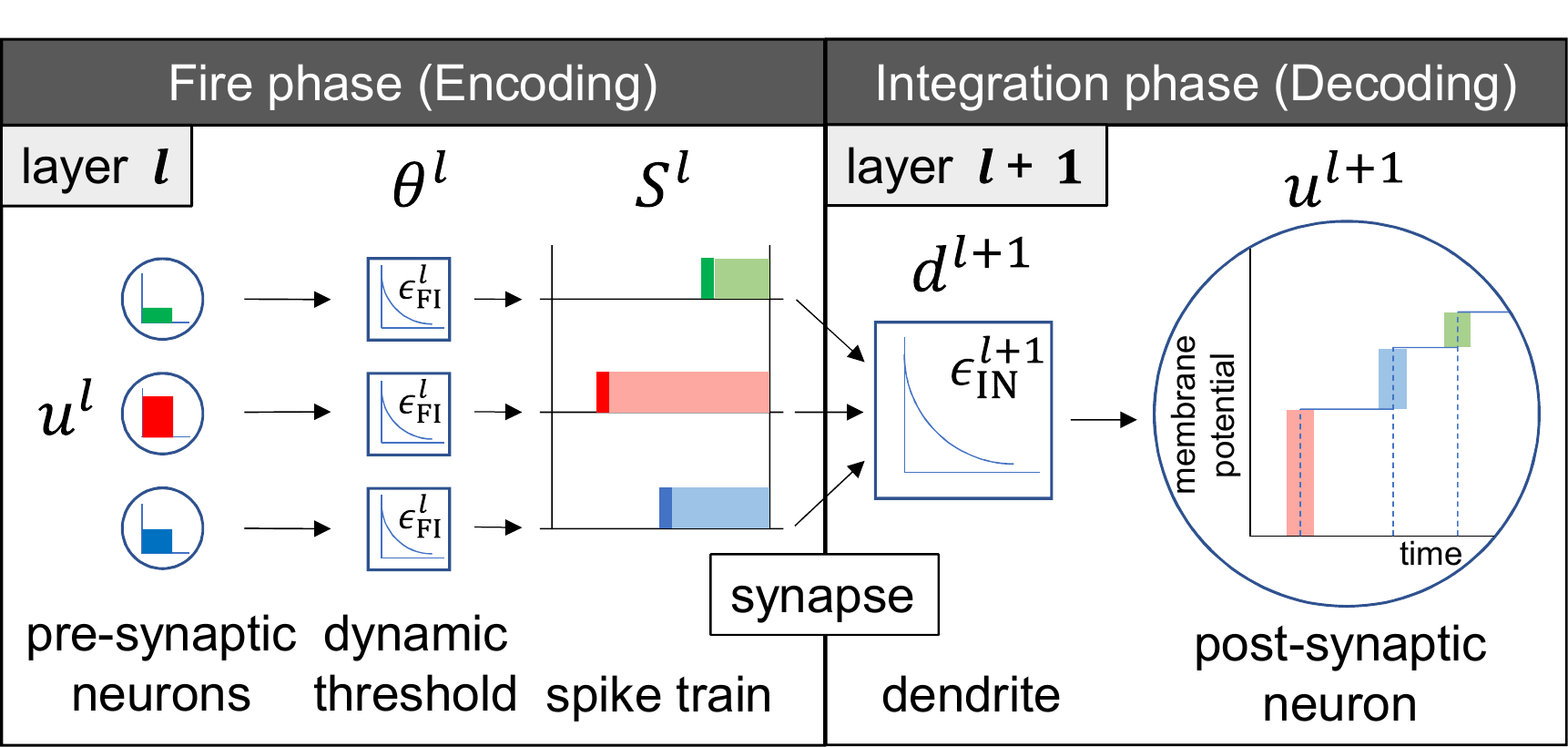}
    \vspace{-0.5em}
	\caption{Overview of T2FSNN: The height of each bars in the presynaptic neurons indicates the amount of integrated membrane potential (information being transmitted).}
	\label{fig:proposed_neural_coding}
    \vspace{-1.75em}
\end{figure}

TTFS coding has been adopted in deep SNNs to overcome such shortcomings~\cite{zhang2019tdsnn,rueckauer2018conversion}.
As illustrated in Fig.~\ref{fig:proposed_neural_coding}, the neurons with TTFS coding generate only one spike during inference and transmit the information using the timing of the spike. 
For instance, a red presynaptic neuron will be the first neuron that generates a spike in the fire phase to encode the largest amount of information (i.e., membrane potential). 
The postsynaptic neuron then integrates the PSP induced by the spike (red bar in $u^{l+1}$).
Note that once a neuron generates a spike, it no longer generates additional spikes by applying a sufficiently long refractory period as described in \cite{zhang2019tdsnn}.

The TDSNN~\cite{zhang2019tdsnn} was able to reduce number of spikes substantially with the reverse coding, which is a kind of TTFS coding.
Reverse coding, however, delivers larger values later, making it difficult to improve latency in deep SNNs.
In addition, the auxiliary neurons, used to implement reverse coding, generate a large number of spikes, which deteriorates the improvement by TTFS coding.
Thus, a new approach to thoroughly utilize the features of TTFS coding is needed for improving the inference of deep SNNs.

\section{Proposed Methods}

In this paper, we propose an SNN model, called T2FSNN, that efficiently implements TTFS coding in deep SNNs.
We adopted the DNN-to-SNN conversion method as described in~\cite{rueckauer2017conversion,kim2019spiking,park2019fast,kim2018deep}.
The T2FSNN exploits the concept of kernel-based dynamic threshold and dendrite for the encoding and decoding procedure, respectively.
In addition, we propose a gradient-based optimization method of the kernels and early firing method for further increasing accuracy and reducing inference latency with the feature of TTFS coding.

\subsection{T2FSNN Model}
The T2FSNN consists of IF neurons, synapses, and dendrites as depicted in Fig.~\ref{fig:proposed_neural_coding}.
In the T2FSNN, each layer has two phases: one is an integration phase (decoding) and the other is a fire phase (encoding) in a time window $T$.
In the integration phase, the neurons in the layer $l+1$ decode the spike trains generated by the presynaptic neurons in the layer $l$, then integrate the decoded information into their membrane potential $u^{l+1}$.
After the integration phase, the integrated membrane potential of a neuron is encoded into a spike. 

The TDSNN~\cite{zhang2019tdsnn} employs TTFS coding which leads to a substantial reduction in number of spikes. 
Nonetheless, performance improvement of TDSNN deteriorated due to the overhead of the auxiliary neurons with a very high firing rate.
In T2FSNN, to efficiently implement TTFS coding without any auxiliary neurons, we designed the encoding and decoding process based on the kernels in the threshold and dendrite, respectively.
The kernels decrease monotonically as
\vspace{-0.4em}
\begin{equation}
\label{eq:kernel}
    \epsilon^{l} (t-t_{\textrm{ref}}^{l})
    = \exp (-({t-t_{\textrm{ref}}^{l} - t_{d}^{l}})/{\tau^{l}}) \textrm{,}
\vspace{-0.4em}
\end{equation}
where $t_{\textrm{ref}}^{l}$ is a reference time that is defined as the start time of the fire phase, $t_{\textrm{d}}^{l}$ is a time delay, and $\tau^{l}$ is a time constant of layer $l$ ($t_{\textrm{d}}^{l}$ and $\tau^{l}$ are trainable parameters of each layer).

The encoding is a process of converting integrated information in the membrane potential $u^{l}$ into a spike time $t^{l}$ at the fire phase of each layer.
We implemented TTFS encoding using a dynamic threshold $\theta^{l}(t)$ described as
\vspace{-0.4em}
\begin{equation}
\label{eq:dynamic_vth}
    \theta^{l}(t) = \theta_{0} \epsilon_{\textrm{FI}}^{l} (t-t_{\textrm{ref}}^{l}) \textrm{,}
\vspace{-0.4em}
\end{equation}
where $\theta_{0}$ is a threshold constant and $\epsilon_{\textrm{FI}}^{l}$ is a fire kernel.
With Eqs.~\ref{eq:firing_time} and~\ref{eq:dynamic_vth}, we were able to obtain the encoding function, which outputs a spike time $t^{l}$ as follows:
\vspace{-0.4em}
\begin{equation}
\label{eq:encoding}
    t^{l} = \lceil -\tau^{l} \ln( u_{i}^{l}(t_{\textrm{ref}}^{l}-1)/\theta_{0}) + t_\textrm{d}^{l} \rceil \textrm{,}
\vspace{-0.4em}
\end{equation}
where $u_{i}^{l}(t_{\textrm{ref}}^{l}-1)$ is the integrated membrane potential of the neuron.
We set the $\theta_0$ to one in this paper, because the range of integrated membrane potentials (activation values in DNNs) was limited [0, 1] by the data-based normalization~\cite{rueckauer2017conversion, park2019fast}.
Because of the characteristic of the dynamic threshold, the larger amount of information integrated, the earlier the membrane potential exceeds the threshold, which means the neuron generates a spike earlier.


The decoding is a process of restoring the information encoded in a spike time at the integration phase.
The decoded information $z_{j}^{l}(t)$ is accumulated in the form of the sum of PSP as follows:
\vspace{-0.4em}
\begin{equation}
\label{eq:decoding}
z_{j}^{l}(t) = \sum\nolimits_{i}{{w_{ij}^{l} \epsilon_{\textrm{IN}}^{l} (t^{l-1}-t_{\textrm{ref}}^{l-1}) }}+b_{j}^{l} \textrm{,}
\vspace{-0.4em}
\end{equation}
where $\epsilon_{\textrm{IN}}^{l}$ is an integration kernel.
To make the decoding more accurate and effective, we set the time constant $\tau^{l}$ and time delay $t_{\textrm{d}}^{l}$ in the integration kernel $\epsilon_{\textrm{IN}}^{l} (t)$ to be equal to those in the fire kernel of the previous layer $\epsilon_{\textrm{FI}}^{l-1} (t)$.

\subsection{Gradient-Based Optimization Method}

The key factors of T2FSNN are the integration and fire kernel in the dendrite and dynamic threshold, respectively.
The kernels have a significant effect on the efficiency of inference in T2FSNN.
The transmission error caused by the kernels mainly consists of two factors: precision error and small value encoding error.
The precision error occurs because the precision of the information to be delivered is different from the precision that the kernels can represent.
We can define the precision error as $|x-\hat{x}|$, where $x$ and $\hat{x}$ are the information before encoding and restored after decoding, respectively.
With Eqs.~\ref{eq:firing_time}, \ref{eq:kernel}, and \ref{eq:encoding}, we can obtain the precision error as $\hat{x}(\exp(1/\tau)-1)$.
The precision error is inversely proportional to $\tau$, which means that the error can be reduced by increasing $\tau$.

However, because the encoding is based on the threshold operation, the information smaller than the smallest value ($\exp(-(T-t_{\textrm{d}})/\tau)$) that the kernel can express in a given time window $T$ cannot be transmitted.
To prevent this loss in transmission of small values, $\tau$ is sufficiently small to represent the values.
Thus, there is a trade-off between the precision and latency of information transmission between neurons, depending on the time constant $\tau$ in the kernel.


To address such a trade-off properly, we propose loss functions, including precision loss and representation loss, considering the accuracy and latency of the inference.
In addition, we propose a gradient-based optimization, which is based on supervised learning and minimizes the loss functions in a layer-wise manner.
We set ground truth of the supervised learning to $\bar{z}$, which is the value from DNN used in the DNN-to-SNN conversion.
The precision loss is defined as 
\vspace{-0.5em}
\begin{equation}
\label{eq:loss_precision}
L_{\textrm{pre}}^{l} = \frac{1}{|F^{l}|}\sum\nolimits_{f \in F^{l}}{\frac{1}{2}(\bar{z}_{f}^{l}-\hat{z}_{f}^{l})^2} \textrm{,}
\vspace{-0.4em}
\end{equation}
where $F^{l}$ is a set of spike time of all neurons in layer $l$, $\bar{z}_{f}^{l}$, and $\hat{z}_{f}^{l}$ are the ground truth in DNN, and the decoded value in T2FSNN corresponds to the spike time $f$, respectively.

The representation loss consists of two terms; $L_{\textrm{min}}^{l}$ and $L_{\textrm{max}}^{l}$.
These two loss terms consider the minimum and maximum representation values of each layer's kernel, so that the kernel can learn the distribution of ground truth $\bar{z}^{l}$. 
$L_{\textrm{min}}^{l}$ is defined as 
\vspace{-0.4em}
\begin{equation}
\label{eq:loss_min}
L_{\textrm{min}}^{l} = {\frac{1}{2}(\bar{z}_{\textrm{min}}^{l}-\hat{z}_{\textrm{min}}^{l})^2} \textrm{,}
\vspace{-0.4em}
\end{equation}
where $\bar{z}_{\textrm{min}}^{l}$ is the minimum value of $\bar{z}^{l}$, and $\hat{z}_{\textrm{min}}^{l}$ is the minimum value that the kernel can represent in a given time window $T$, which is $\exp(-(T-t_{\textrm{d}}^{l})/\tau^{l})$.
$L_{\textrm{max}}^{l}$ is defined as 
\vspace{-0.4em}
\begin{equation}
\label{eq:loss_max}
L_{\textrm{max}}^{l}= {\frac{1}{2}(\bar{z}_{\textrm{max}}^{l}-\hat{z}_{\textrm{max}}^{l})^2} \textrm{,}
\vspace{-0.4em}
\end{equation}
where $\bar{z}_{\textrm{max}}^{l}$ is the maximum value of $\bar{z}^{l}$, and $\hat{z}_{\textrm{max}}^{l}$ is the maximum value that the kernel can represent, which is $\exp(t_{\textrm{d}}^{l}/\tau^{l})$.

\begin{figure}[tbp]
    \centering
    \includegraphics[width=0.9\linewidth]{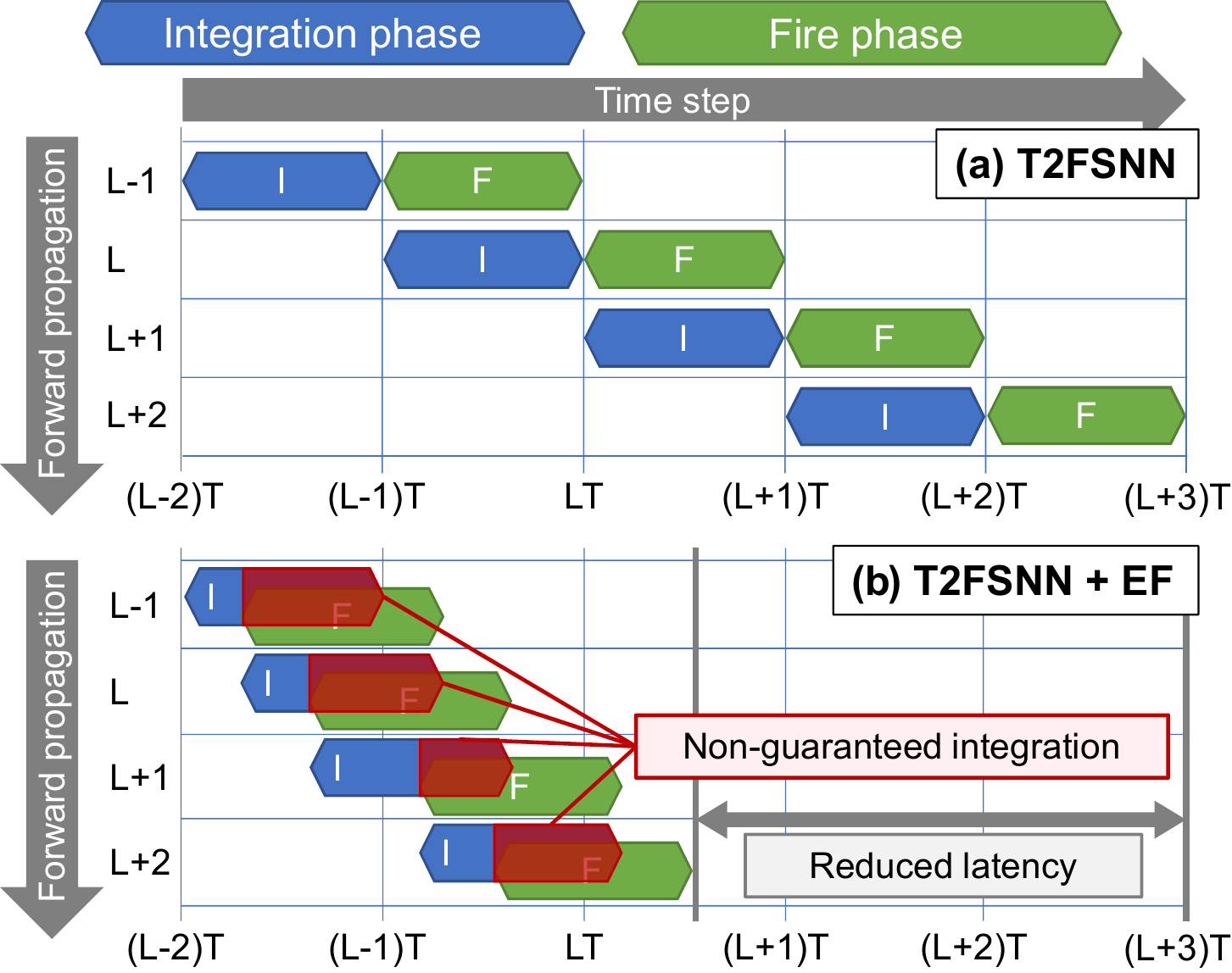}
	\caption{Pipeline of the integration and fire phase}
	\label{fig:early_firing}
    \vspace{-2.0em}
\end{figure}


To minimize the proposed loss functions, we use a gradient-based optimization method. 
The derivative of the precision loss with respect to $\tau^{l}$ is obtained as
\vspace{-0.3em}
\begin{equation}
\label{eq:grad_loss_precision}
\frac{\partial{L_{\textrm{prec}}^{l}}}{\partial{\tau}^{l}}
= \frac{\partial{L_{\textrm{prec}}^{l}}}{\partial{\hat{z}^{l}}} \frac{\partial{\hat{z}^{l}}}{\partial{\tau^{l}}}
= -\frac{1}{|F^{l}|}\sum\nolimits_{f \in F^{l}}{\frac{t_{f}-t_{\textrm{d}}^{l}}{(\tau^{l})^{2}}(\bar{z}_{f}^{l}-\hat{z}_{f}^{l})\hat{z}_{f}^{l}}
\vspace{-0.4em}
\end{equation}
and the derivative of minimum representation loss with respect to $\tau^{l}$ is stated as
\vspace{-0.4em}
\begin{equation}
\label{eq:grad_loss_min}
\frac{\partial{L_{\textrm{min}}^{l}}}{\partial{\tau^{l}}}
= \frac{\partial{L_{\textrm{min}}^{l}}}{\partial{\hat{z}_{\textrm{min}}^{l}}} \frac{\partial{\hat{z}_{\textrm{min}}^{l}}}{\partial{\tau^{l}}}
= -{\frac{T-t_{\textrm{d}}^{l}}{(\tau^{l})^2}({z}_{\textrm{min}}^{l}-\hat{z}_{\textrm{min}}^{l})\hat{z}_{\textrm{min}}^{l}} \textrm{.}
\vspace{-0.4em}
\end{equation}
Because the maximum representation is most affected by $t_{\textrm{d}}$, maximum representation loss is differentiated with respect to $t_{\textrm{d}}$ as follows:
\vspace{-0.4em}
\begin{equation}
\label{eq:grad_loss_max}
\frac{\partial{L_{\textrm{max}}^{l}}}{\partial{t_{\textrm{d}}^{l}}}
= \frac{\partial{L_{\textrm{max}}^{l}}}{\partial{\hat{z}_{\textrm{max}}^{l}}} \frac{\partial{\hat{z}_{\textrm{max}}^{l}}}{\partial{t_{\textrm{d}}^{l}}}
= -\frac{1}{\tau^{l}}({z}_{\textrm{max}}^{l}-\hat{z}_{\textrm{max}}^{l})\hat{z}_{\textrm{max}}^{l} \textrm{.}
\vspace{-0.4em}
\end{equation}
From the equations above, we can calculate the gradient of $\tau^{l}$ and $t_{\textrm{d}}^{l}$, and optimize the kernel function with those gradients.

\subsection{Early Firing Method}

\begin{figure}
    \centering
    \includegraphics[width=0.9\linewidth]{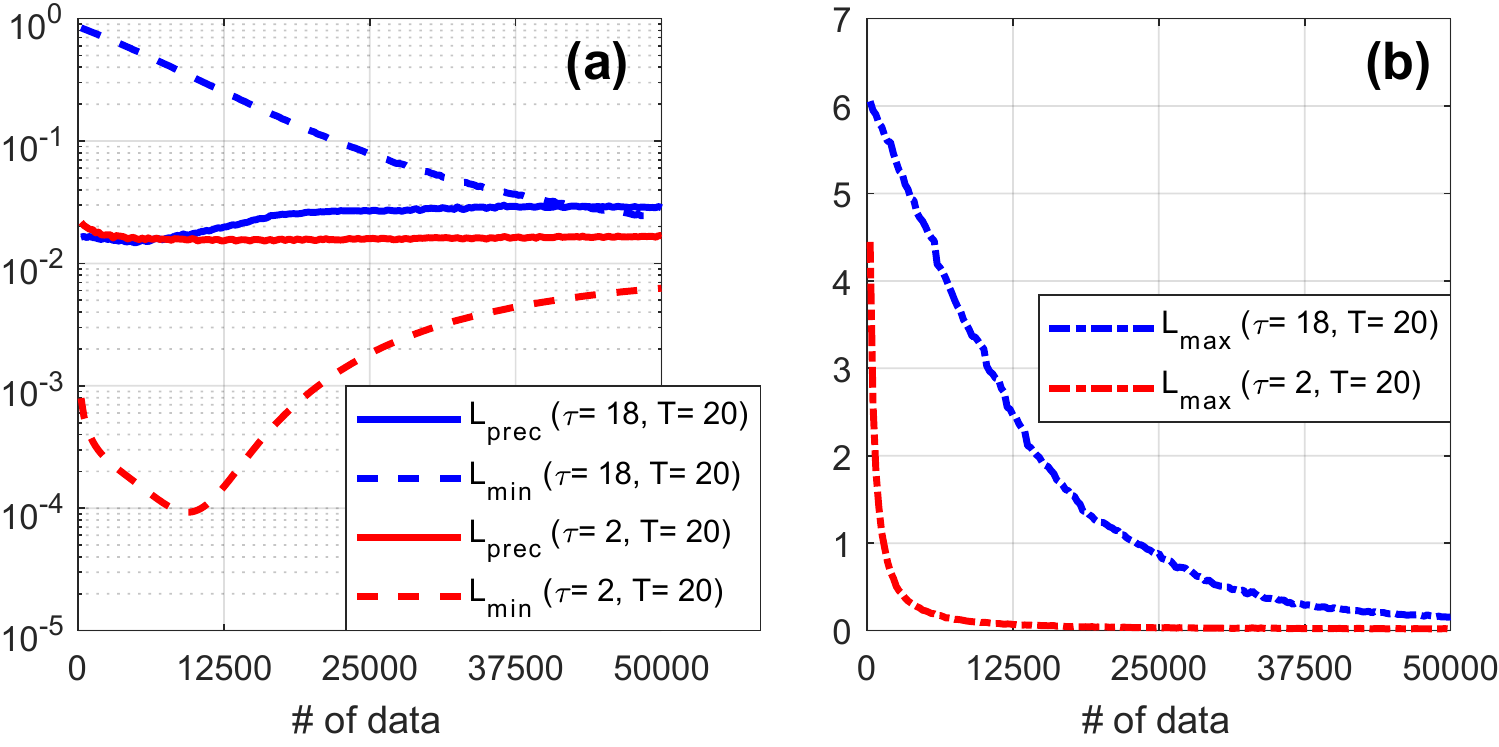}
    \vspace{-0.5em}
    \caption{Loss graphs of the proposed gradient-based optimization method: (a) $L_{\textrm{prec}}$ and $L_{\textrm{min}}$, (b) $L_{\textrm{max}}$}
    \label{fig:loss_optimization}
    \vspace{-1.5em}
\end{figure}



The integration and fire phase of T2FSNN are operated with dependencies in order, as shown in Fig.~\ref{fig:early_firing}.
In each layer of this baseline pipeline (Fig.~\ref{fig:early_firing}-(a), T2FSNN), the integration and fire phases in a layer are executed sequentially.
The integration phase of layer $l$ is executed simultaneously with the fire phase of the previous layer $l-1$ from $(L-1)T$ to $LT$ time step.
After the integration is finished, the fire phase begins at $LT$ time step, generating spikes based on the integrated information.
This baseline pipeline can guarantee the integration of all information from the previous layer in a given time window $T$ before encoding, which leads to accurate information transmission to the subsequent layer.
However, the dependency between integration and fire phase considerably increases inference latency in deep SNNs with many layers.

To alleviate this issue, we propose a technique called early firing, which is a method of starting the firing phase before the integration is complete at each layer.
By using this method, we can overlap the integration and fire phase, which results in reducing inference latency as shown in Fig.~\ref{fig:early_firing}-(b) (T2FSNN+EF).
However, this approach causes non-guaranteed integration, where the information may not contribute to generating a spike.
The information accumulated in the neurons that have already fired during the non-guaranteed integration cannot affect the spike generation, because each neuron generates at most one spike in the T2FSNN.
Thus, the starting time of the early firing should be set to ensure the guaranteed integration of critical information to reduce inference latency without significant loss of accuracy.

\section{Experimental Results}
We empirically set the $\tau$, $t_{\textrm{d}}$, and $T$ at the initial stage.
Based on the initial configurations of T2FSNN, we applied the proposed gradient-based optimization and early firing.
We used a train dataset and mini-batch SGD for the optimization.
For the early firing, we set the starting time of the early firing to half of the time window $T$ based on the experiments.
We assessed the T2FSNN, including the proposed methods, and compared the other neural coding methods on various datasets.

\subsection{Evaluation of the Proposed Methods}

\begin{figure}[tbp]
    \centering
    \includegraphics[width=0.9\linewidth]{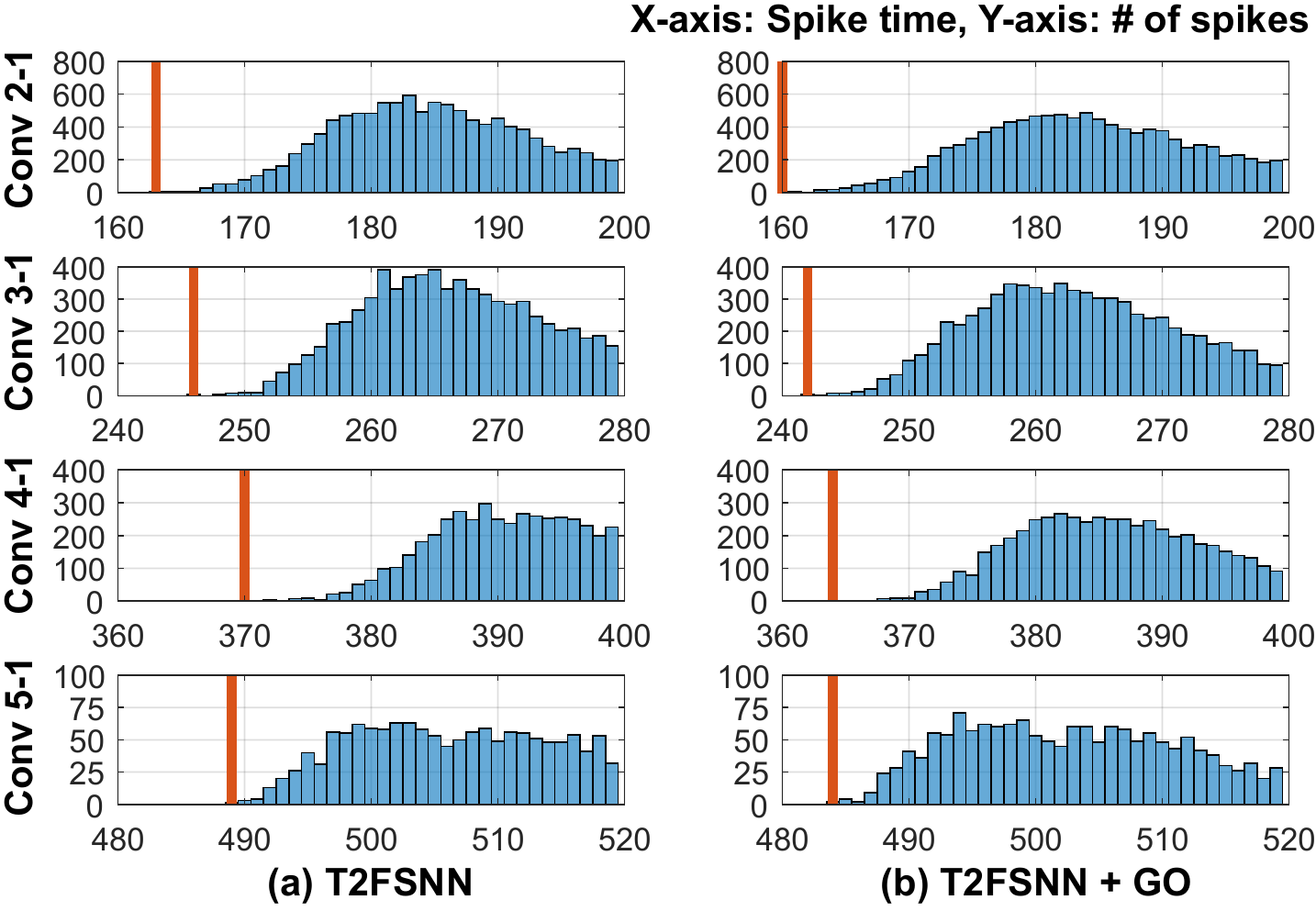}
    \vspace{-0.5em}
	\caption{Distributions of spike time: Vertical orange bars represent the first spike time of each layer (VGG-16, CIFAR-10).}
	\label{fig:dist_spike_time}
    \vspace{-1.5em}
\end{figure}

To evaluate the proposed optimization method, we measured the three loss terms ($L_{\textrm{prec}}$, $L_{\textrm{min}}$, and $L_{\textrm{max}}$).
We set two different initial conditions to validate the trade-off between precision and latency of information transmission depending on the $\tau$: 1) a small time constant ($\tau$=2), and 2) a large time constant ($\tau$=18) on a given time window ($T$=20).
When the time constant is a small value ($\tau$=2), the kernel $\epsilon$ can represent small values sufficiently, but the precision of transmission is low.
Thus, as the training progresses, the time constant $\tau$ increased and the precision loss $L_{\textrm{prec}}$ decreased as shown in Fig.~\ref{fig:loss_optimization}-(a) (red solid line). 
In contrast, if the time constant is a large value ($\tau$=18), the proposed method will train $\tau$ to minimize the minimum representation loss $L_{\textrm{min}}$ as seen in the blue dashed line on Fig.~\ref{fig:loss_optimization}-(a), which results in decreasing $\tau$.

The maximum representation loss is depicted in Fig.~\ref{fig:loss_optimization}-(b).
In the case of a small time constant ($\tau$=2), the maximum representation loss $L_{\textrm{max}}$ decreases rapidly because the first spike of each layer occurs at an earlier time in each layer's fire phase. 
Through this experiment, we can verify that $L_{\textrm{prec}}$ and $L_{\textrm{min}}$ are trained competitively in a given time window $T$, and $L_{\textrm{min}}$ has a greater impact than $L_{\textrm{prec}}$.
We can also validate the effect of the gradient-based optimization by the distribution of spike time in each layer as depicted in Fig.~\ref{fig:dist_spike_time}. 
Compared to the T2FSNN (Fig.~\ref{fig:dist_spike_time}-(a)), the optimized T2FSNN (T2FSNN+GO, Fig.~\ref{fig:dist_spike_time}-(b)) can shorten the first spike time of each layer (vertical orange bar), and reduce the number of spikes.
This indicates that the T2FSNN+GO enables the efficient inference that is fast and requires less computation.

\setlength{\textfloatsep}{0pt}
\ctable[
pos= t,
center,
caption = {Ablation study},
captionskip = -1.0ex,
label = {tab:ablation_study},
doinside = {\footnotesize \def\arraystretch{1.0} \setlength{\tabcolsep}{2.3pt}}
]{lccccc}{
    \tnote[a]{GO: Gradient-based Optimization, EF: Early Firing}
}
{
    \toprule
    \multirow{2}{*}{Methods} & \multirow{2}{*}{Latency } & \multicolumn{2}{c}{CIFAR-10} & \multicolumn{2}{c}{CIFAR-100}\\ 
    \cmidrule(r){3-4}
    \cmidrule(r){5-6}
    & & Accuracy & Spikes & Accuracy & Spikes\\
    \midrule
    T2FSNN & 1280 & 91.36 & 6.898E+4 & 66.04 & 8.626E+4\\
    T2FSNN+GO\tmark[a] & 1280 & 91.37 & 6.887E+4 & 66.97 & 8.464E+4\\
    T2FSNN+EF\tmark[a] & 680 & 91.37 & 6.893E+4 & 68.09 & 8.603E+4\\
    \midrule
    \textbf{T2FSNN+GO+EF}\tmark[a] & \textbf{680} & \textbf{91.43} & \textbf{6.881E+4} & \textbf{68.79} & \textbf{8.444E+4}\\
	\bottomrule
}

We conducted an ablation study to demonstrate individual effects of the proposed methods as described in Table~\ref{tab:ablation_study}.
When the gradient-based optimization was applied to the T2FSNN (T2FSNN+GO), accuracy on CIFAR-10 and CIFAR-100 improved by approximately 0.01\% and 0.93\% compared to the T2FSNN, respectively. 
In addition, the number of spikes decreased by 0.2\% and 1.9\%. 
When the early firing method was applied to the T2FSNN (T2FSNN+EF), inference latency decreased by 46.9\%, while accuracy increased by 2.05\% on CIFAR-100. 
Remarkably, the T2FSNN+GO+EF achieved 46.9\% reduction in inference latency while improving accuracy of 0.07\% and 2.75\% on CIFAR-10 and CIFAR-100, respectively. 
Furthermore, the number of spikes also decreased by 0.3\% and 2.1\%. 
It is interesting to note that the accuracy was improved with a lower number of spikes, despite that early firing causes non-guaranteed integration.
This result can be interpreted as a generalization effect caused by the stochastic property of non-guaranteed integration that contributes to the generation of a spike depending on the state of the neuron.

\subsection{Comparison with Other Methods}

\begin{figure}[tbp]
    \centering
    \includegraphics[width=0.91\linewidth]{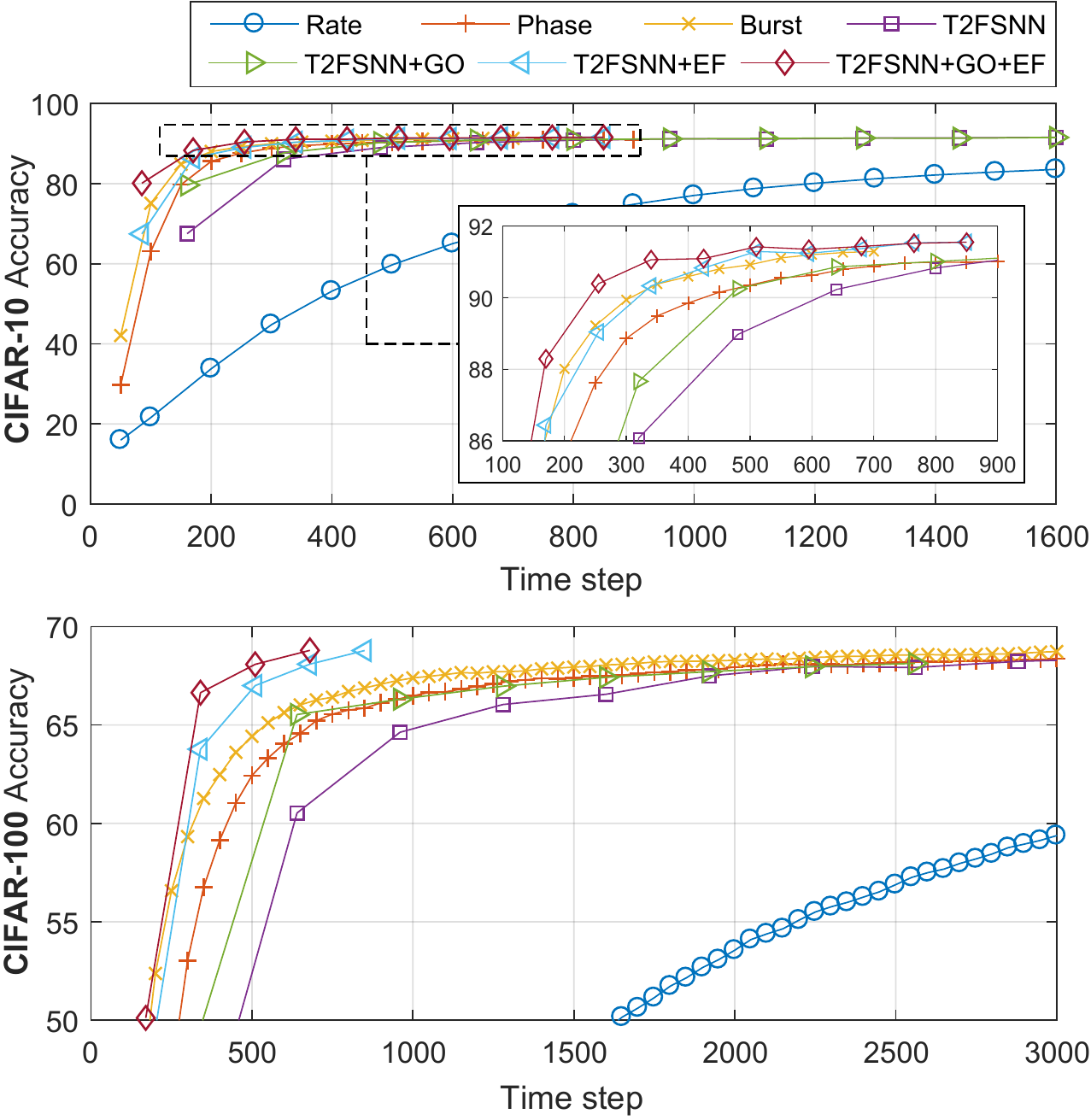}
	\vspace{-0.5em}
	\caption{Inference curve of various neural coding methods for VGG-16 on the CIFAR-10 (top) and CIFAR-100 (bottom)}
	\label{fig:inference_curve}
\end{figure}

We compared our proposed approach with the other neural coding methods, including rate~\cite{diehl2015fast,rueckauer2017conversion}, phase~\cite{kim2018deep}, burst coding~\cite{park2019fast}.
To assess inference speed, we measured the accuracy depending on the inference time as depicted in Fig.~\ref{fig:inference_curve}.
In CIFAR-10, the order of fast-to-slow inference speed is burst, phase, T2FSNN, and rate coding. 
When we applied the proposed optimization or early firing to the T2FSNN (T2FSNN+GO, T2FSNN+EF), the inference speed of T2FSNN+GO and T2FSNN+EF was as fast as that of phase and burst coding, respectively.
In the case of T2FSNN with the optimization and early firing, T2FSNN+GO+EF showed the fastest inference speed.
We observed similar tendency in the experimental results on CIFAR-100.
T2FSNN+EF showed faster inference speed than burst coding, and T2FSNN+GO+EF achieved the fastest inference speed. 

\setlength{\textfloatsep}{0pt}
\ctable[
pos = t,
center,
caption = {Comparison with other DNN-to-SNN conversion methods with various neural coding schemes},
captionskip = -1.0ex,
label = {tab:experimental_result_other},
doinside = {\footnotesize \def\arraystretch{.8} \setlength{\tabcolsep}{5pt}}
]{l|crr|rr}{
}{
    \toprule
    Neural & Accuracy & \multicolumn{1}{c}{Latency} & \multicolumn{1}{c|}{Spikes} & \multicolumn{2}{c}{Normalized Energy}\\
    Coding & (\%) & \multicolumn{1}{c}{(time step)} & \multicolumn{1}{c|}{($10^6$)} & \multicolumn{1}{c}{TN~\cite{merolla2014million}} & \multicolumn{1}{c}{SN~\cite{furber2014spinnaker}} \\
	\midrule
	\multicolumn{6}{l}{MNIST} \\
	\midrule
	Rate~\cite{diehl2015fast,rueckauer2017conversion} & 99.10 & 200 & 0.100 & 1.000 & 1.000 \\
	Phase~\cite{kim2018deep} & 99.20 & \textbf{16} & 3.000 & 12.048 & 19.228 \\
	Burst~\cite{park2019fast} & 99.25 & 87 & 0.251 & 1.265 & 1.763 \\
	Reverse~\cite{zhang2019tdsnn} & 99.08 & - & - & -  & - \\
	\textbf{Our Method} & \textbf{99.33} & 40 & \textbf{0.002} & \textbf{0.128} & \textbf{0.085} \\

	\midrule
	\multicolumn{6}{l}{CIFAR-10} \\
	\midrule
	Rate~\cite{diehl2015fast,rueckauer2017conversion} & 91.14 & 10,000 & 61.949 & 1.000 & 1.000 \\
	Phase~\cite{kim2018deep} & 91.21 & 1,500 & 35.196 & 0.317 & 0.418 \\
	Burst~\cite{park2019fast} & 91.41 & 1,125 & 6.920 & 0.112 & 0.112 \\
	\textbf{Our Method} & \textbf{91.43} & \textbf{680} & \textbf{0.069} & \textbf{0.041} & \textbf{0.025} \\
	
	\midrule
	\multicolumn{6}{l}{CIFAR-100} \\
	\midrule
	Rate~\cite{diehl2015fast,rueckauer2017conversion} & 66.50 & 10,000 & 81.525 & 1.000 & 1.000 \\
	Phase~\cite{kim2018deep} & 68.66 & 8,950 & 258.408 & 1.805 & 2.351 \\
	Burst~\cite{park2019fast} & 68.77 & 3,100 & 25.074 & 0.309 & 0.308 \\
	\textbf{Our Method} & \textbf{68.79} & \textbf{680} & \textbf{0.084} & \textbf{0.041} & \textbf{0.025} \\
	
	\bottomrule
}

Table~\ref{tab:experimental_result_other} summarizes the accuracy, latency, number of spikes, and estimated energy consumption of various neural coding on MNIST, CIFAR-10, and CIFAR-100. 
Our method (T2FSNN+GO+EF) achieved the best accuracy with the lowest number of spikes in all datasets.
Particularly, in CIFAR-100, we were able to reduce the number of spikes to less than 1\% of that of burst coding, given the fact that the T2FSNN generates at most one spike per neuron. 
Inference latency also decreased to 22\% compared to that of burst coding.

The energy consumption takes both latency and number of spikes into account, and thus it is an important metric for the efficiency of deep SNNs~\cite{park2019fast}.
We estimated the energy consumption based on measured latency and spikes with dynamic and static energy consumption data from neuromorphic architecture, TrueNorth (TN)~\cite{merolla2014million} and SpiNNaker (SN)~\cite{furber2014spinnaker} as in \cite{park2019fast}.
The estimated energy is defined as $(\textrm{\# of spikes})E_{\textrm{dyn}} + (\textrm{latency})E_{\textrm{sta}} \textrm{,}$ where $E_{\textrm{dyn}}$ and $E_{\textrm{sta}}$ are dynamic and static energy parameters depeneding on neuromorphic architecture.
The energy parameters ($E_{\textrm{dyn}}$, $E_{\textrm{sta}}$) are set to (0.4, 0.6) and (0.64, 0.36) for TrueNorth and SpiNNaker, respectively.
According to the estimation, our method was able to reduce energy consumption to about 6\% and 16\% on average compared to rate and burst coding, respectively.

\section{Discussion}

The TDSNN (reverse coding), which is a previous study in the line of our work, did not report the number of spikes and latency~\cite{zhang2019tdsnn}.
Instead of direct comparison to TDSNN, we compared the estimated computational cost as stated in Table~\ref{tab:computational_cost}.
The rate coding only requires accumulation of incoming spikes.
The phase and burst coding, on the other hand, need additional non-linear functions during encoding and decoding procedure.
Such extra overheads are alleviated by using a lookup table due to the limited input range of the non-linear function.
Thus, these neural coding methods require multiplication and addition operations proportional to the number of spikes.

The TDSNN used auxiliary neurons called ticking neurons for implementing reverse coding.
The ticking neurons generate spikes frequently, which causes numerous accumulation operations.
In addition, the TDSNN adopted leaky IF neurons, which require an exponential operation at every time step.
Thus, required computations are proportional to the time step and number of neurons.
The estimated computational cost based on the reported data~\cite{zhang2019tdsnn} is stated in Table~\ref{tab:computational_cost}. 
In contrast, the proposed T2FSNN does not require auxiliary neurons, the computational cost of kernel function in T2FSNN can be reduced by replacing the kernel with a lookup table.
According to our analysis, the T2FSNN requires significantly lower operations compared to various neural coding schemes including TDSNN.

\section{Conclusion}

\setlength{\textfloatsep}{0pt}
\ctable[
pos=t,
center,
caption = {Analysis of computational cost (million operations, VGG-16 on CIFAR-100)},
captionskip = -1.0ex,
label = {tab:computational_cost},
doinside = {\footnotesize \def\arraystretch{1.0} \setlength{\tabcolsep}{5pt}}
]{lcccccc}{
}{
    \toprule
    & \multirow{2}{*}{DNN} & Rate & Phase & Burst & TDSNN & \multirow{2}{*}{\textbf{T2FSNN}} \\
    & & \cite{diehl2015fast,rueckauer2017conversion} & \cite{kim2018deep} & \cite{park2019fast} & \cite{zhang2019tdsnn} & \\
    \midrule
    Mult & 146.50 & - & 258.408 & 25.074 & 14.84 & \textbf{0.084}\\
    Add & 146.50 & 81.525 & 258.408 & 25.074 & 154.21 & \textbf{0.084}\\
	\bottomrule
}

In this paper, we proposed the T2FSNN with gradient-based optimization and early firing for utilizing the TTFS coding in deep SNNs.
By the extensive experiments on various tasks, we demonstrated our methods improve the inference efficiency of deep SNNs in terms of both latency and number of spikes.
We expect that our methods pave the way for energy-efficient inference with deep SNNs with temporal coding.


%

\section*{Acknowledgments}
\begin{small}
This work was supported in part by the National Research Foundation of Korea (NRF) grant funded by the Korea government (Ministry of Science and ICT) [2016M3A7B4911115, 2018R1A2B3001628], the Brain Korea 21 Plus Project in 2019, AIR Lab (AI Research Lab) in Hyundai Motor Company through HMC-SNU AI Consortium Fund, and Samsung Research Funding \& Incubation Center of Samsung Electronics under Project Number SRFC-IT1901-12.





\end{small}

\bibliographystyle{IEEEtran}
\bibliography{dac2020}

\end{document}